\documentclass{article}

\usepackage{arxiv}

\usepackage[utf8]{inputenc} 
\usepackage[T1]{fontenc}    
\usepackage{hyperref}       
\usepackage{url}            
\usepackage{booktabs}       
\usepackage{amsfonts}       
\usepackage{nicefrac}       
\usepackage{microtype}      
\usepackage{lipsum}
\usepackage{eurosym}

\usepackage{balance} 
\usepackage{amsmath}
\usepackage{mathrsfs}
\usepackage{bm}
\usepackage{booktabs}
\usepackage{amsfonts,amssymb}
\usepackage{colortbl}
\usepackage{color}
\usepackage{tikz}
\usepackage{multicol}
\usepackage{threeparttable}
\usepackage{makecell}
\usepackage{multirow}

\usepackage{algorithmicx}
\usepackage{algpseudocode}
\usepackage[]{algorithm2e}
\usepackage{siunitx}
\usepackage{gensymb}

\usepackage{graphicx}

\usepackage{upgreek}
\usepackage{amsmath,amssymb,amsfonts}

\usepackage{amsmath}

\title{Lo-MARVE: A Low Cost Autonomous Underwater Vehicle for Marine Exploration}

\author{
  Karl Mason\\
  School of Computer Science\\ 
  University of Galway\\
  Galway, Ireland\\
  \texttt{karl.mason@universityofgalway.ie} \\
   \And
 Daniel Kelly \\
  School of Engineering\\ 
  University of Galway\\ 
  Galway, Ireland\\
  \texttt{daniel.kelly@universityofgalway.ie} \\
}

\begin{document}
\maketitle

\begin{abstract}
This paper presents Low-cost Marine Autonomous Robotic Vehicle Explorer (Lo-MARVE), a novel autonomous underwater vehicle (AUV) designed to provide a low cost solution for underwater exploration and environmental monitoring in shallow water environments. Lo-MARVE offers a cost-effective alternative to existing AUVs, featuring a modular design, low-cost sensors, and wireless communication capabilities. The total cost of Lo-MARVE is approximately EUR 500. Lo-MARVE is developed using the Raspberry Pi 4B microprocessor, with control software written in Python. The proposed AUV was validated through field testing outside of a laboratory setting, in the freshwater environment of the River Corrib in Galway, Ireland. This demonstrates its ability to navigate autonomously, collect data, and communicate effectively outside of a controlled laboratory setting. The successful deployment of Lo-MARVE in a real-world environment validates its proof of concept.
\end{abstract}




\section{Introduction}\label{sec:Intro}

\let\thefootnote\relax\footnotetext{\textit{Proc. of the 12th International Conference on Control, Mechatronics and Automation (ICCMA 2024), London, UK, November 11-13, 2024, \url{https://www.iccma.org/}. 2024.}}

Autonomous Underwater Vehicles (AUVs) have emerged as indispensable tools for exploring the vast and largely unexplored depths of the oceans. Their ability to operate independently, navigate complex underwater environments, and collect valuable data has revolutionized fields such as marine science, oceanography, and resource exploration. AUVs are equipped with advanced sensors, communication systems, and propulsion mechanisms, enabling them to perform tasks that would be impractical or dangerous for human divers. The development of robust and reliable AUV technologies is crucial for addressing pressing global challenges, including climate change, marine conservation, and sustainable resource management. 


Advanced AUVs, such as Vityaz-D, have reached depths of over 10km \cite{alekseev2021alternative}. These highly capable AUVs offer advanced capabilities and can reach significant depths. The main limitation of these advanced AUVs is that they can be prohibitively expensive. Purchasing such AUVs presents a barrier to researchers with modest budgets who wish to conduct research on AUVs. This provides the motivation for research into low cost AUVs.

There have been many low cost AUVs published in the literature. For example LoCo was proposed in 2020 \cite{edgeDesignExperimentsLoCO2020}. Studies such as this demonstrate that it is possible to develop capable AUVs for lower cost. The proposed AUV aims to provide a more accessible and affordable option for researchers and organizations seeking to conduct underwater exploration and data collection. The main advantage of the proposed AUV is its low cost when compared to existing low cost AUVs.


The contributions of this paper are:
\begin{itemize}
    \item The design and development of a very low cost AUV for marine robotics research. 
    \item To deploy the proposed AUV in the field to demonstrate it's functionality.
\end{itemize}

The rest of the paper is structured as follows. Section 2 will give an overview of the research in the literature in low cost AUV development. Section 3 will outline the design of the proposed ultra low cost AUV. Section 4 will then outline the experimental procedure for evaluating the AUV. The results of AUV testing are presented in Section 5. Finally, Section 6 will draw conclusions from this research.

\section{Related Work}
\label{sec:relatedW}
One of greatest contributors to cost in AUVs is the size of the unit, as material and electronics requirements increase with the scale of the vessel. Therefore, the most low-cost solutions are in the small-AUV class of vessel (typically under one meter in length).  Commercial offerings such as the Bluefin Robotics Sandshark, HII REMUS 100 and Teledyne Gavia offer reliable performance but are high-cost investments in the tens of thousands of dollars. These vessels use a modular design to reduce cost from their larger counterparts. Following this ethos there are number of proposed low-cost designs in academic literature with similar capabilities. 

Ribas et al. who introduce the Girona 500 AUV which has a multi-hull design for switching between survey and intervention tasks, offering a greater value vessel \cite{ribas2011girona}.
Zhou et al. developed ALPHA, a torpedo style AUV featuring tunnel thruster propulsion for acrobatic underwater movements. The system is integrated with Robot Operating System (ROS) and is fully simulated in the Stonefish simulation environment. The cost of manufacturing ALPHA was under 14,000 USD \cite{zhouAcrobaticLowcostPortable2022}. 

Ridolfi et al. for the euRathlon 2015 developed FeelHippo, a Remotely Operated Vehicle (ROV) style vessel that can also operate as an AUV for approximately 13,000 EUR \cite{ridolfiFeelHippoLowcostAutonomous2016}.
Edge et al. propose LoCO, also a ROS integrated and simulated in Gazebo modular vessel. It is a dual torpedo style AUV that uses computer vision for localisation and command following. It has a lower build cost of 4000 USD \cite{edgeDesignExperimentsLoCO2020}. 

Conversely to a modular design, mission specific AUV design has also proven to be beneficial to reducing manufacturing costs. Specifically for the task of shallow water ($<30m$ depth) surveying, unique configuration can be used that offer better performance of the vessel as well as lower build cost.

Alvarez et al. specifically mention this approach in their development of Fólaga. A torpedo surface gliding AUV that uses ballast to dive and hold station \cite{alvarezFolagaLowcostAutonomous2009}. Dunbanin et al. address the task of coral reef navigation with their AUV Starbug. It uses computer vision for path finding through the informationally dense landscape and extra control surfaces for a high degree of manoeuvrability. Their system build cost 10,000 AUS \cite{dunbabinHybridAUVDesign2005}. While modular in its hardware configuration, Duecker et al. present an extremely low cost design, HippoCampusX, a micro class AUV that is specifically rated for shallow water, multi-bot missions. The cost of each unit is 800 USD \cite{dueckerHippoCampusXHydrobaticOpensource2020}. For further reading, Sanchez et al. provide a more comprehensive review of the state of art in novel and main instrumentation and measurement systems \cite{sanchez2020autonomous}.

\section{Lo-MARVE Design}
\label{sec:Lo-MARVE}

\begin{figure}[!t]
\centering
\includegraphics[width=3.2in]{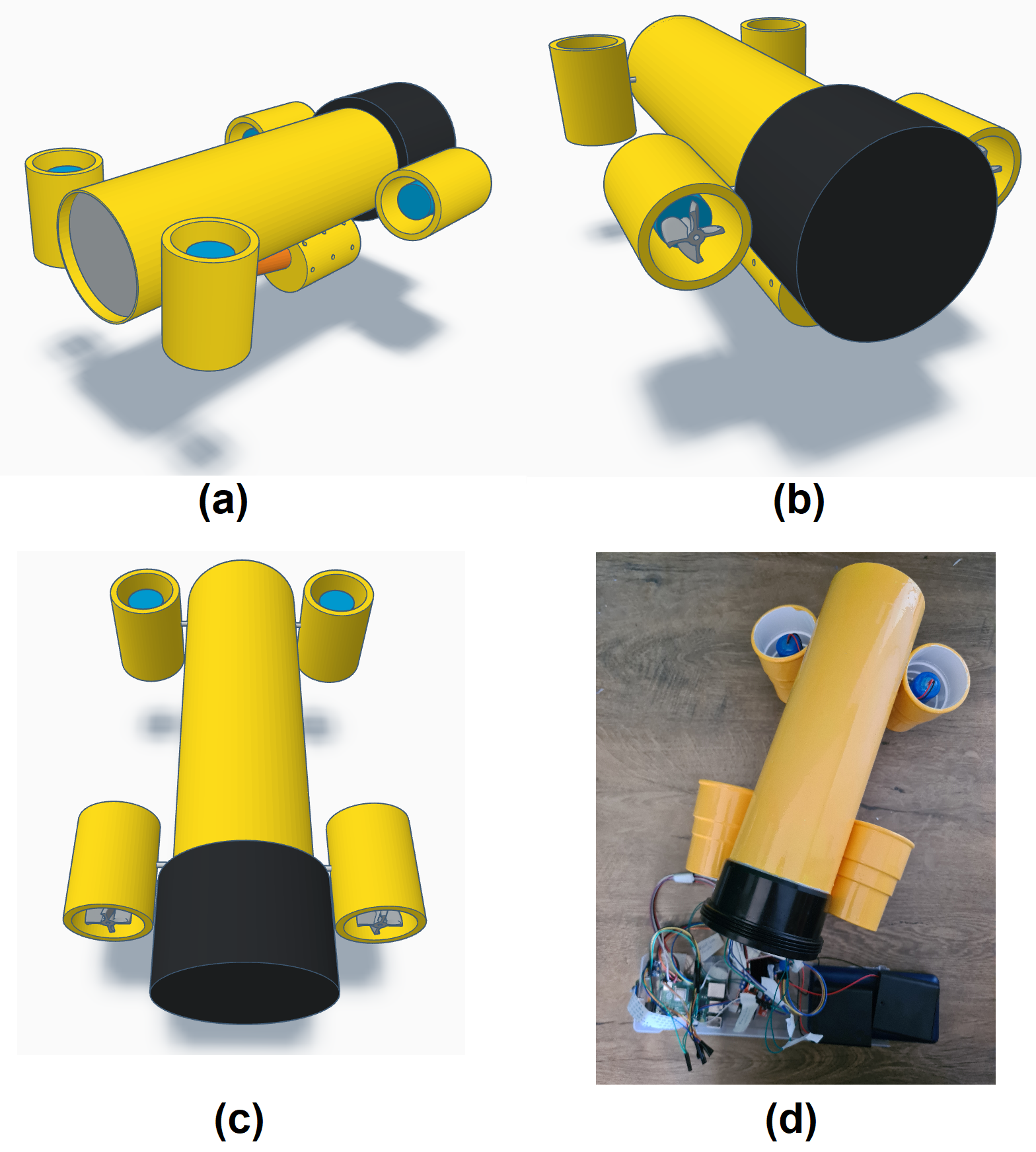}
\caption{(a), (b) and (c) CAD design of Lo-MARVE. (d) Final build of Lo-MARVE.}
\label{fig:AUV}
\end{figure}

The design of the Lo-MARVE is depicted in Figure \ref{fig:AUV}. 

\begin{figure}[!h]
\centering
\includegraphics[width=3.2in]{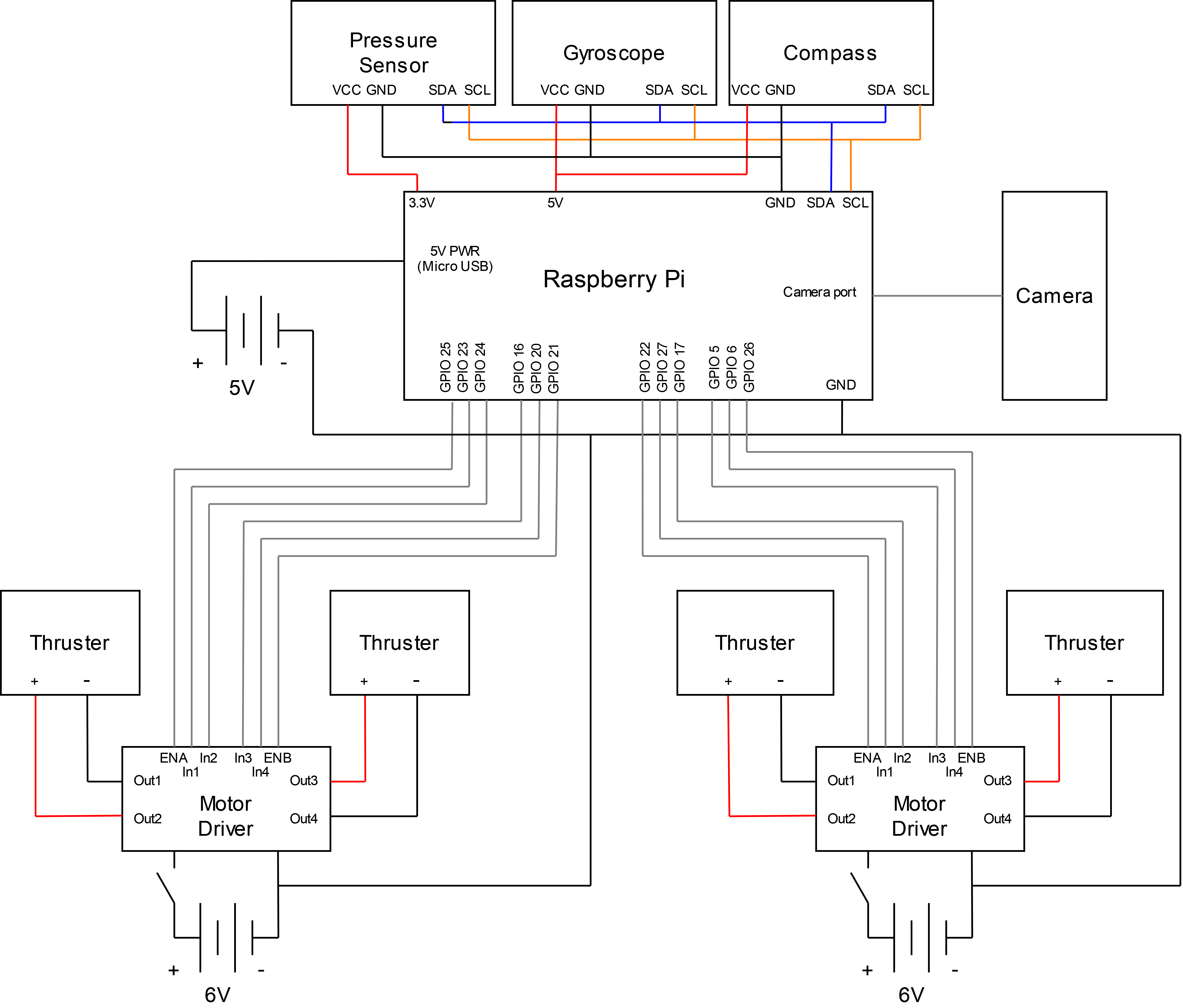}
\caption{Lo-MARVE Wiring Diagram.}
\label{fig:circuit}
\end{figure}

The Lo-MARVE AUV primarily consists of a single cylindrical hull for electronics, sensors and batteries. This PVC compartment has a diameter of $100mm$ and length $300mm$. The nose is enclosed using a transparent acrylic disc sealed using epoxy resin. The rear of the AUV fixed with a removable end cap to provide access to the electronics. The end cap section has a larger diameter of $130mm$ and extends the total length of the AUV to $380mm$. Figure \ref{fig:circuit} outlines the wiring of the Lo-MARVE AUV.

\subsection{Motors}
A total of 4 motors are mounted to the AUV, 2 at the nose of the AUV and 2 at the back of the AUV. The 2 motors mounted at the front are orientated to provide thrust such that the pitch of the AUV can be adjusted. The orientation of the 2 rear motors is such that they provide forward/backward thrust. The motors used for the AUV were 16800RPM Lichifit motor thrusters with an operating voltage of 3V-12V. The motors are fixed to PVC covers for protection, which are mounted to the craft using stainless steel bolts. 

\subsection{Microprocessor and Sensors}
The AUV is controlled using the Raspberry Pi microprocessor. The control software for the AUV is written in Python. The microprocessor is powered using a 20 Ah power bank. The sensors connected to the microprocessor are:
\begin{itemize}
    \item Camera - Raspberry Pi Camera Module 2
    \item Compass - GY-271 (qmc5883l) compass
    \item Gyroscope - GY-512 (MPU-6050) gyroscope
    \item Pressure sensor - Adafruit LPS35HW sensor
\end{itemize}

The camera is mounted within the AUV at the nose in order to record the underwater environment through the transparent window. The compass and gyroscope are mounted next to the camera within the AUV. These are used to track the orientation of the AUV. 
The pressure sensor is mounted outside of the AUV in order to measure depth. The pressure sensor is enclosed within a rubber balloon that is protected by a PVC cover. This is a similar approach to that proposed by MIT Sea Grant\cite{MITSeaGrant}. The pressure sensor is connected to the microprocessor via a rubber hose. The pressure sensor is mounted under the main compartment of the AUV.

    	   	




\subsection{Control Software}


The main control loop of the AUV is outlined in Algorithm \ref{Alg:Control}. All Python classes and sensors are initialized at the beginning of the programme. At each loop iteration, the controller executes a command, e.g. calibrate the robot sensors, descend to the target depth and move forward for a specified time, etc. Once the command has been executed, the bluetooth connection is re-established and a new command is requested from the user.

\begin{algorithm}[]
\caption{Main control loop of AUV}
\begin {algorithmic}
\State \textbf{Initialize} AUV, sensors and communications
\State \textbf{Calibrate} sensors
\State \textbf{Set} experimental parameters
\State \textbf{Test} bluetooth connection
\State \textbf{Get} command from user\\
\While{command $!=$ end}{
    \State Execute command
    \State Re-establish bluetooth connection
    \State Get new command
}
\State \textbf{Write} data to file
\State \textbf{Terminate} bluetooth connection
\State \textbf{Shutdown} microprocessor\\
\end{algorithmic}
\label{Alg:Control}
\end{algorithm}

The pseudocode outlined in Algorithm \ref{Alg:forward} describes the controller that enables the Lo-MARVE AUV moves forward. This is an example of routine that will be executed when a command is sent to the robot from within the control loop in Algorithm \ref{Alg:Control}.

\begin{algorithm}[]
\caption{Forward Movement}
\begin {algorithmic}
\State \textbf{Initialize} sensors and get initial readings
\State \textbf{Set} target heading
\State \textbf{Record} start time
\State \textbf{Move forward}\\
\While{current time $<$ end time}{
    \State Get depth and headingError\\
    \If{pitch too steep}{
        \State Adjust pitch, move forward
    }\\
    \ElseIf{too deep or too shallow}{
        \State Ascend/Descend, move forward
    }\\
    \ElseIf{heading off-target}{
        \State Adjust heading, move forward
    }\\
    \Else{
        \State Continue moving forward
    }
    \State Get new sensor readings
}
\State \textbf{Stop} AUV\\
\end{algorithmic}
\label{Alg:forward}
\end{algorithm}

        

\subsection{Communications}

\begin{figure}[h]
\centering
\includegraphics[width=3.5in]{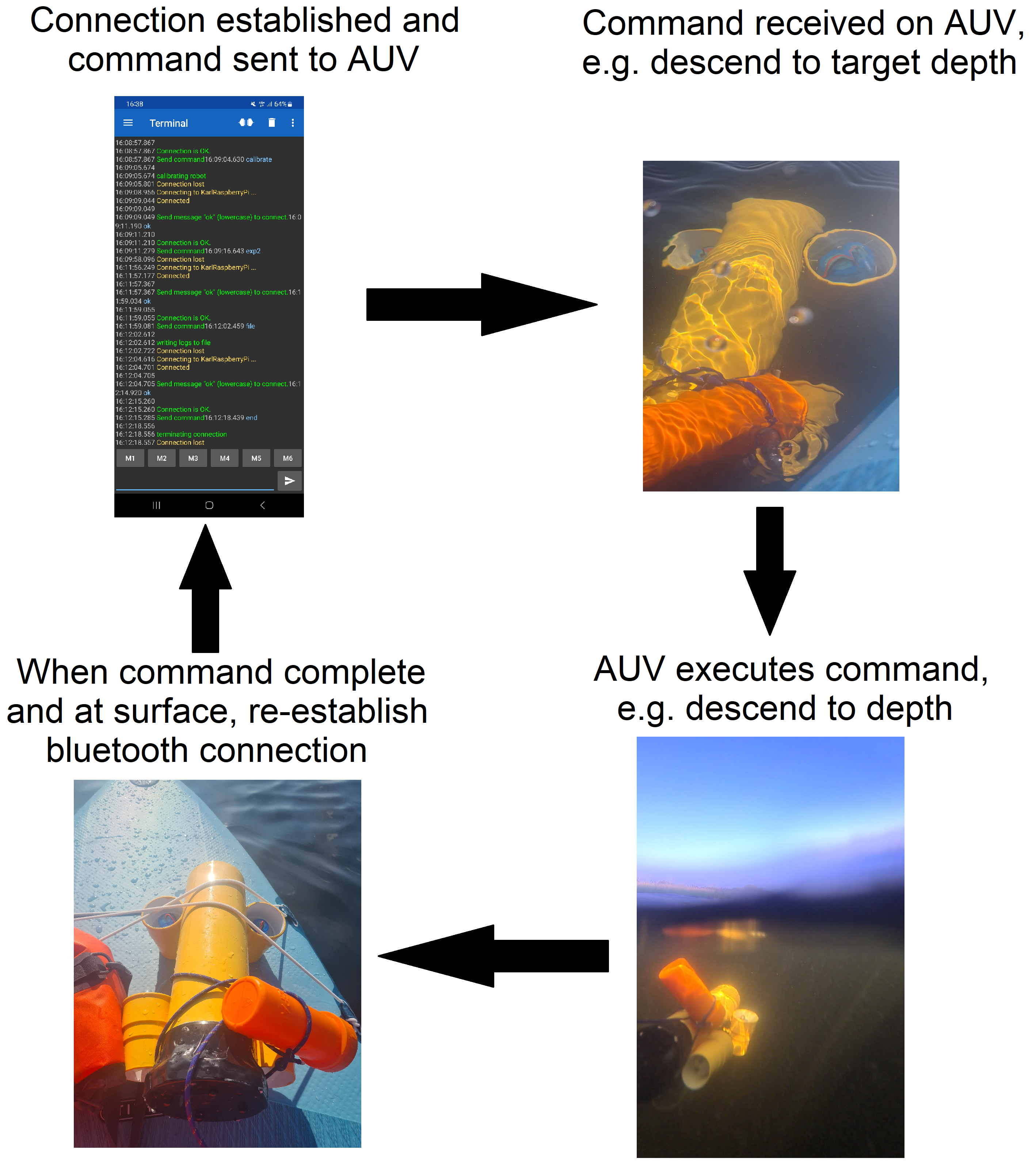}
\caption{Bluetooth Communications with Lo-MARVE.}
\label{fig:bluetooth}
\end{figure}

The AUV communicates with the user via bluetooth when the AUV is at the surface. Commands are sent to the AUV using the Serial Bluetooth Terminal application on an android device. This is illustrated in Figure \ref{fig:bluetooth}. When the AUV resurfaces, the bluetooth connection is re-established.


\section{Experimental Setup}
\label{sec:Experiments}


\begin{table}[!b]
\centering
\caption{Summary of Lo-MARVE Parameters}
\scalebox{1.0}{
\begin{tabular}{c | c }
\noalign{\smallskip}\hline \noalign{\smallskip}
    
        Parameter   	&	Details 	 \\

	\noalign{\smallskip}\hline\noalign{\smallskip}
Total hull length	    &	$380$ mm	\\
Hull internal diameter	    &  	$100$ mm	\\
Volume (excluding additional cylinder)	    &  	$0.00395$ $m^3$	\\
Mass (excluding added weights)	    &  	$2.351$ kg	\\
Battery capacity    &   20 Ah \\
Microprocessor	    &  	Raspberry Pi 4b	\\
Controller programming language	    &  	Python	\\

\noalign{\smallskip}
\hline
\end{tabular}} 
\label{table:param}
\end{table}

\begin{figure*}[!t]
\centering
\includegraphics[width=6in]{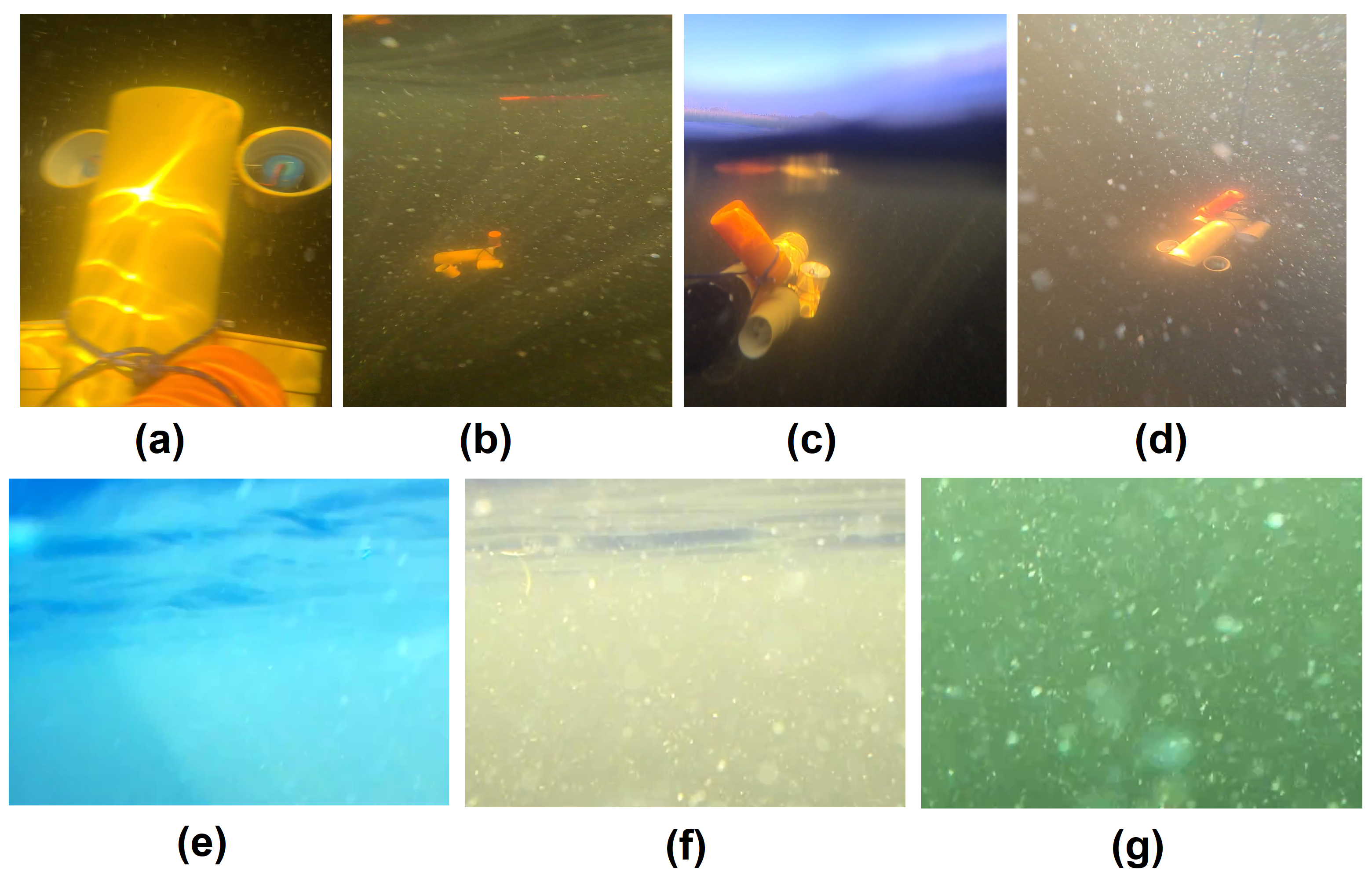}
\caption{Lo-MARVE Underwater Field Test. (a), (b), (c), (d) are images captured by experimenter. (e), (f) and (g) are images captured by Lo-MARVE.}
\label{fig:underwater}
\end{figure*}

In order to achieve neutral buoyancy, the volume of Lo-MARVE was measured by measuring the volume of displaced water after submerging it in a container of water. The volume of Lo-MARVE is approximately $0.00395 m^3$. The mass of the hull, electronics and batteries of Lo-MARVE is measured to be $2.351 kg$. To achieve neutral buoyancy, $1.599 kg$ of weight was added within the hull, such that the total mass of Lo-MARVE was $3.95kg$. When deployed, a small cylinder is attached to the AUV. This facilitates minor changes in the buoyancy of Lo-MARVE as required. The parameters of Lo-MARVE are summarised in Table \ref{table:param}.

    



For safety, the AUV is attached to flotation device via a rope. The length of the rope is $1.2m$ from the top of the AUV to the flotation device. This rope also attaches the small cylinder to the hull of Lo-MARVE. The flotation device allows the AUV to be easily located when submerged underwater. This flotation device is not required for the AUV to operate. To analyse the trajectory of the AUV, a Garmin Fenix 7 was attached to the flotation device. This device records GPS data. The Lo-MARVE does not receive any GPS data.

The Lo-MARVE was tested in the River Corrib, in Galway, Ireland. This environment was selected as it was a freshwater environment and the weight added to the Lo-MARVE was selected to be neutrally buoyant in freshwater. In addition to this, the location within the River Corrib was selected as there is no fast flowing water or challenging surface conditions. Finally, the depth of the water within the selected location in the River Corrib is relatively shallow. This provides safe conditions to test the Lo-MARVE.

The testing of Lo-MARVE consisted of sending a command via bluetooth to execute two routines:
\begin{enumerate}
    \item Descend to a depth of $1m$.
    \item Continue forward maintaining a constant depth (Algorithm \ref{Alg:forward}).
\end{enumerate}
Note: the depth sensor is mounted underneath Lo-MARVE. Lo-MARVE is also programmed to correct its depth when it is outside of the acceptable deviation from the target depth, i.e $1m \pm 0.25m$. This deviation is specified to prevent the pitch of the AUV from oscillating rapidly around the $1m$ target depth.

\begin{figure}[!h]
\centering
\includegraphics[width=3.0in]{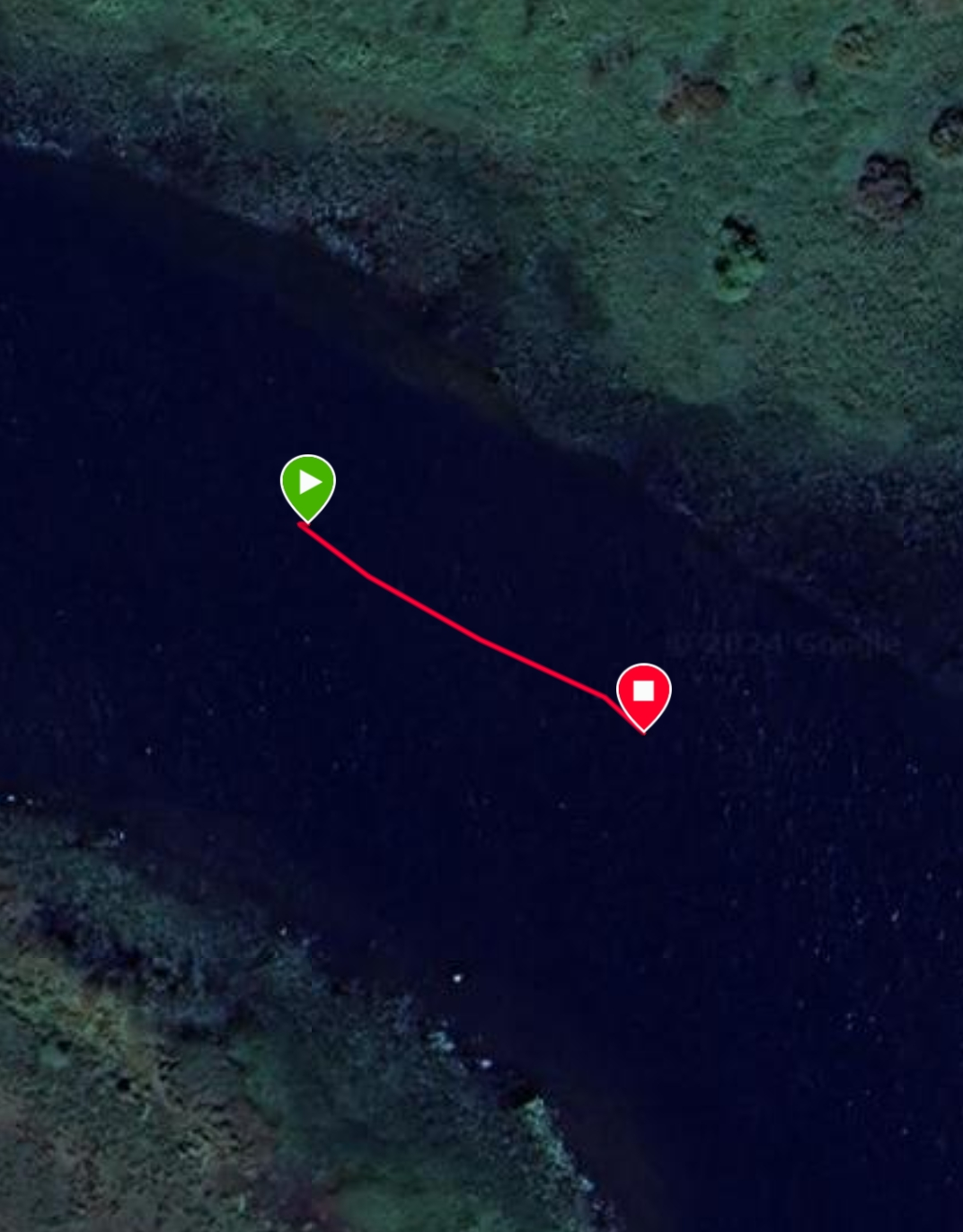}
\caption{GPS data for Lo-MARVE field test.}
\label{fig:GPS}
\end{figure}

\begin{figure*}[!h]
\centering
\includegraphics[width=6.5in]{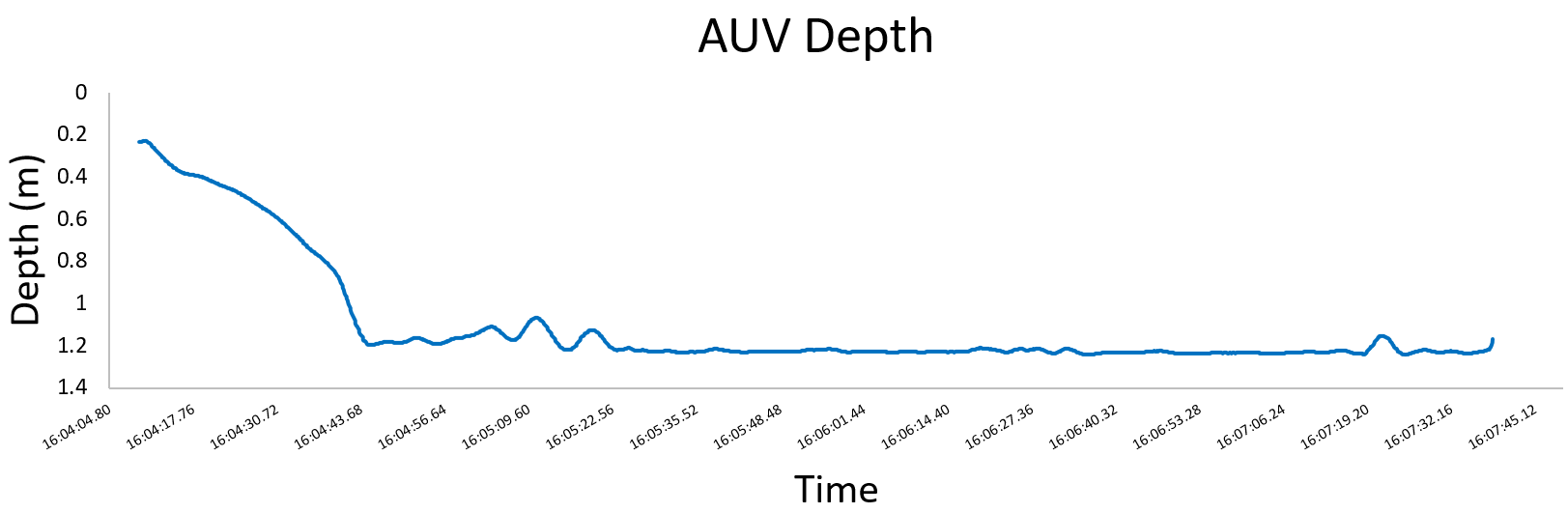}
\caption{Depth of Lo-MARVE.}
\label{fig:depth}
\end{figure*}

\section{Results}
\label{sec:Results}

Figure \ref{fig:underwater} illustrates the Lo-MARVE as it navigates the River Corrib underwater. Figures \ref{fig:underwater} (a), (b), (c) and (d) show Lo-MARVE from the perspective of the experimenter, while Figures \ref{fig:underwater} (e), (f) and (g) depict what is recorded by the Lo-MARVE on-board camera. In Figures \ref{fig:underwater} (a) and (c), Lo-MARVE is close to the surface. Figures \ref{fig:underwater} (b) and (d) show Lo-MARVE when it is at its target depth. Visibility is limited due to the presence of sediment in the river. This is particularly evident in Figures (e), (f) and (g). A change in lighting can be observed in these images. This is caused by changes in depth, orientation, shadows, etc. 







The GPS data for the Lo-MARVE field test presented in Figure \ref{fig:GPS}. This image outlines the path taken by the Lo-MARVE while submerged underwater. The total distance travelled measured by the GPS was $36$ m.




The depth reached by the Lo-MARVE is presented in Figure \ref{fig:depth}. The plot shows the AUV descend to a depth of $\approx 1.2m$. It should be noted that the pressure sensor is mounted beneath the hull of the Lo-MARVE. Therefore when the AUV is at the surface, a depth reading of $\approx 0.2$ m is recorded. This plot shows how the Lo-MARVE descends to the target depth within 1 minute. The AUV approaches the lower bound of the acceptable depth $1m \pm 0.25m$. Figure \ref{fig:depth} then shows that the depth recorded oscillates briefly before leveling out. The total span time of the x axis of Figure \ref{fig:depth} is slightly under $4$ minutes.



\subsection{Discussion}
\label{sec:discussion}
The results presented in this section demonstrate the viability of the proposed Lo-MARVE. The Lo-MARVE is capable of operating while fully submerged in a freshwater environment outside of a laboratory setting. Lo-MARVE can capture video data from the underwater environment for later analysis.

The primary advantage of the proposed Lo-MARVE is it's low cost. The Lo-MARVE costs $\approx$\euro$500$ in total. This can facilitate researchers with a moderate budget to conduct research on AUVs. This research builds on existing studies published that propose low cost AUVS, as outlined in Section \ref{sec:relatedW}. For example, one of the lowest cost AUVs in the literature was the HippoCampusX, proposed by Duecker et al. for $\$800$\cite{dueckerHippoCampusXHydrobaticOpensource2020}. The Lo-MARVE presented in this paper can rival this AUV in terms of cost.

One example of Lo-MARVE low cost design features is the PVC cylinder and end cap utilized for the Lo-MARVE hull. This has a cost of $\approx$ \euro $20$ - \euro $40$. Commercially available watertight tubes that are designed for use as AUV hulls can cost $\$230 - \$330$. These more expensive hulls have been used in other low cost AUVs in the literature \cite{edgeDesignExperimentsLoCO2020}.

The low cost parts used for the proposed Lo-MARVE are also it's limitation. The thrusters used for Lo-MARVE were selected for their low cost ($\approx$\euro$60$ for 4 Lichifit motor thrusters). These motors do not have the same durability or power as more expensive thrusters used in other AUV designs. Reduced durability is also a limitation for other components in the Lo-MARVE design, e.g. the PVC hull. Commercially available hulls have a specified depth rating of $>100$ m. The maximum depth of the Lo-MARVE has not been tested, however it is expected to be far more limited than other commercial watertight enclosures.

Notwithstanding these limitations, Lo-MARVE presents a low cost AUV design with fewer capabilities than more expensive alternatives. AUV design can be framed as a multi-objective optimisation problem where there is a trade-off between the minimizing the cost objective and the maximizing the performance/functionality objective. Lo-MARVE, and other AUVs in the literature \cite{dueckerHippoCampusXHydrobaticOpensource2020,edgeDesignExperimentsLoCO2020} place a high weighting to the cost objective.

    

	


\section{Conclusion}
\label{sec:Conclusion}
This paper proposes a low cost AUV,  the Low-cost Marine Autonomous Robotic Vehicle Explorer (Lo-MARVE). This AUV is designed to provide a very low cost alternative to other AUVs available on the market. The hull, motors, sensors, and electronics were all selected to minimize cost. The Lo-MARVE was tested in an open body of freshwater, specifically the River Corrib in Galway, Ireland. This validates the design of the Lo-MARVE AUV.

\subsection{Future Work}
\label{sec:future}
There are several avenues for future work that have arisen from this research. These include:
\begin{itemize}
    \item Testing Lo-MARVE in a saltwater environment. This will require adjusting the weight of Lo-MARVE to account for the saltwater environment. Enabling Lo-MARVE to effectively operate in a saltwater environment would expand the use cases for Lo-MARVE.
    \item Upgrading the thrusters/batteries to enhance the capability of Lo-MARVE. Currently, Lo-MARVE struggles to navigate against the current in moving water. Exploring more effective thrusters would help to overcome this limitation. Of course, this would likely increase the build cost of Lo-MARVE.
    \item Enhancing the control systems of Lo-MARVE, e.g. to navigate an environment in a grid search type pattern. This would further establish the capabilities of Lo-MARVE.   
\end{itemize}





\bibliographystyle{ACM-Reference-Format} 
\bibliography{references}

\end{document}